\documentclass[conference]{IEEEtran}
\IEEEoverridecommandlockouts

\def\BibTeX{{\rm B\kern-.05em{\sc i\kern-.025em b}\kern-.08em
    T\kern-.1667em\lower.7ex\hbox{E}\kern-.125emX}}
\newcommand{\etal}{\textit{et al.}}
\newcommand{\cmmnt}[1]{}

\usepackage{multirow}
\usepackage{subcaption} 
\usepackage{amsmath,amssymb,amsfonts}

\usepackage{xcolor}
\newcommand{\ak}[1]{\textcolor{black}{#1}}

\usepackage[most]{tcolorbox}
\usepackage{enumitem}

\usepackage{newtxtext}
\usepackage[cal=boondox]{mathalfa}

\DeclareMathAlphabet{\pazocal}{OMS}{zplm}{m}{n}   

\usepackage{cite}
\usepackage{amsmath,amssymb,amsfonts}
\usepackage{algorithmic}
\usepackage{graphicx}
\usepackage{xcolor}
\usepackage{comment}
\usepackage{multirow}
\usepackage{booktabs}
\usepackage{fancyhdr}
\usepackage{mathtools}

\begin{document}


\title{Split-Fuse-Transport: Annotation-Free Saliency via Dual Clustering and Optimal Transport Alignment}


\author{
   \small
   Muhammad Umer Ramzan\textsuperscript{1}, Ali Zia\textsuperscript{2}, Abdelwahed Khamis\textsuperscript{3}, Noman Ali\textsuperscript{1}, Usman Ali\textsuperscript{1}, Wei Xiang\textsuperscript{2} \\
   \textsuperscript{1}GIFT University, Pakistan  \textsuperscript{2}La Trobe University, Australia
   \textsuperscript{3}CSIRO, Australia \\
   \{umer.ramzan, usmanali, 211980037\}@gift.edu.pk, \{A.Zia, W.Xiang\}@latrobe.edu.au, abdelwahed.khamis@data61.csiro.au 
}

\maketitle



\begin{abstract}
Salient object detection (SOD) aims to segment visually prominent regions in images and serves as a foundational task for various computer vision applications.\ak{We posit that SOD can now reach near-supervised accuracy without a single pixel-level label, but only when reliable pseudo-masks are available.
We revisit the prototype-based line of work and make two key observations.
First, boundary pixels and interior pixels obey markedly different geometry, second, the global consistency enforced by optimal transport (OT) is underutilised if prototype quality is weak.
}\ak{To address this, we introduce POTNet, an adaptation of Prototypical Optimal Transport that replaces POT’s single k-means step with an entropy-guided dual-clustering head: high-entropy pixels are organised by spectral clustering, low-entropy pixels by k-means, and the two prototype sets are subsequently aligned by OT.  This split–fuse–transport design yields sharper, part-aware pseudo-masks in a single forward pass, without handcrafted priors. Those masks supervise a standard MaskFormer‐style encoder–decoder, giving rise to AutoSOD, an end-to-end unsupervised SOD pipeline that eliminates SelfMask’s offline voting yet improves both accuracy and training efficiency. Extensive experiments on five benchmarks show that AutoSOD outperforms unsupervised methods by up to 26\% and weakly supervised methods by up to 36\% in F-measure, further narrowing the gap to fully supervised models.
}

\end{abstract}


\section{Introduction}
Salient object detection (SOD) is the task of segmenting an image at the pixel level to highlight regions that are likely to attract human attention. It plays a key role as a prior in various computer vision applications, such as image segmentation, 
image captioning,
and object tracking ~\cite{gurkan2021tdiot}. Earlier approaches relied on hand-crafted features, including colour histograms, boundary connectivity, and high-dimensional colour transformations. However, these methods often fail to generate high-quality saliency maps in cluttered scenes, particularly when foreground objects share visual similarities with the background. In recent years, deep convolutional neural networks (CNNs) have led to substantial improvements in both image segmentation and salient object detection by enabling more robust and context aware feature representation. However, the scalability of such a supervised learning approach is limited, as collecting ground truth annotations is a labour intensive and costly process.

To reduce the need for manual labelling, weakly supervised approaches \cite{zeng2019multi} have been introduced. These methods tackle saliency detection using sparse supervisory signals, such as image-level class labels \cite{li2018weakly} or descriptive captions \cite{zeng2019multi}. It is important to note that \cite{zeng2019multi} depends on features or attention maps obtained from classification tasks, which are trained using global image-level labels or captions. As a result, they may struggle to generate high-quality pseudo ground truths.

To reduce the reliance on large-scale human annotations, numerous unsupervised methods for saliency detection and object segmentation have been proposed in recent years \cite{zhang2020learning, piao2021mfnet, yan2022unsupervised, wang2022multi}. Nevertheless, a considerable performance gap still exists between unsupervised and fully supervised SOD approaches.

In the salient object detection (SOD) task, Class Activation Maps (CAMs) \cite{zhou2016learning} are widely used to localise objects and to generate pseudo labels that subsequently train a SOD model \cite{chen2017deeplab}. However, CAMs tend to emphasise only the most discriminative foreground regions \cite{zhang2023credible}, offering limited supervision for SOD task training \cite{chen2017deeplab}. To alleviate this issue, prototype‑based methods \cite{zhao2024sfc, zhao2024psdpm} have been introduced. These approaches first derive binary masks for each class using feature similarity \cite{zhao2024sfc}, and then the masks are used via mask average pooling on the corresponding features, yielding a prototype for every class in an image. These prototypes capture the essential characteristics of foreground objects. 
Yet a single prototype strategy is still unable to capture salient objects efficiently as features closest to the prototype are strongly activated, while more distant regions receive only partial activation. This incomplete coverage provides inadequate guidance during SOD task and ultimately limits performance.

\ak{Pixel features in unsupervised saliency detection are commonly grouped by \textit{k}-means\cite{gupta2021aw}, whose objective minimises within–cluster variance and therefore favours \textit{compact}, roughly spherical groups \cite{MacQueen1967}.  This compactness is ideal for the many \textbf{low-entropy} pixels that already form dense, homogeneous blobs (e.g., clear sky or pavement).  However, the same Euclidean, isotropic partitioning fragments \textbf{high-entropy} pixels—those with ambiguous CAM scores that lie on curved or intertwined manifolds (e.g. a flower’s spiralling petals, or ring-shaped wheels)—either splitting these parts or merging them with the background.  
Spectral methods inspired by Normalised Cut \cite{Shi2000} overcome this by cutting along affinity-graph boundaries and have recently delivered state-of-the-art unsupervised saliency; SelfMask’s spectral-cluster voting \cite{Shin2022} and TokenCut’s transformer-token graph cuts \cite{Wang2022TokenCut} both show sizeable gains over \emph{k}-means baselines.  
Yet applying spectral clustering indiscriminately can over-segment compact regions and adds significant computational burden.}

\ak{We therefore propose an \textbf{entropy-guided hybrid clustering} that \emph{bridges} these complementary strengths.  Pixels whose CAM entropy exceeds a threshold are routed to spectral clustering to respect global structure while the remaining low-entropy pixels are clustered by \emph{k}-means to retain local compactness.  
A soft entropy-based coefficient fuses the two assignments into unified memberships, from which multiple prototypes are averaged.  
These graph-aware prototypes are then fed to the Similarity-Aware Optimal Transport module of POT \cite{Wang2025POT}, aligning the full feature distribution with the prototype distribution under global consistency constraints.  
This selective spectral–\textit{k}-means fusion and its seamless integration with OT constitute a key methodological contribution of our work, advancing beyond pipelines that rely solely on \emph{k}-means or global spectral cuts without transport-level alignment.
}

The main contributions of our proposed framework are outlined as follows:
\begin{enumerate}

\item \ak{\textbf{Dual-clustering prototype generator (POTNet):}
We adapt POT \cite{Wang2025POT} to saliency by splitting pixels with a CAM-entropy gate: spectral clustering on high-entropy (boundary) pixels and k-means on low-entropy (interior) pixels.  The two prototype sets are then aligned through Optimal Transport, producing sharp, part-aware pseudo-masks without human labels.}

\item \ak{\textbf{One-Pass Mask Supervision:}  POTNet delivers one-pass pseudo-masks that supervise a DETR-style encoder–decoder during training, eliminating the offline pre-processing step that SelfMask \cite{Shin2022} and other unsupervised SOD pipelines require. Thus, cutting the training time and providing stronger supervision.}


\item \ak{\textbf{SOTA Unsupervised Saliency:} The combined system outperforms previous unsupervised and weakly supervised methods on five public datasets, narrowing the gap to fully supervised saliency detectors—all with lower computational cost.}

\end{enumerate}

\section{Related Work}\label{sec:related_work}

Traditional SOD methods rely on hand-crafted features and heuristics, including template matching, colour histograms, which perform well in simple settings but fail in complex scenes with cluttered backgrounds, blurred boundaries, or irregular object shapes~\cite{hou2018deeply}.

Recent advances in deep learning have led to the development of various supervised CNN-based methods \cite{liu2018picanet, wu2019mutual,qin2020u2, liu2021visual} that significantly enhance SOD performance. These approaches commonly utilise strategies such as multi-scale contextual feature extraction \cite{10869577} and boundary-aware modelling. Zhao \etal~\cite{zhao2019egnet} integrated salient edge and region information to improve boundary precision; however, this approach remains sensitive to cluttered and low-contrast environments. Qin \etal~\cite{qin2019basnet} introduced dual U-Net architectures for iterative refinement of saliency maps, albeit with increased computational overhead. GateNet~\cite{zhao2020suppress} employed a gated dual-branch structure \cite{10.1007/978-981-96-5223-5_33} to strengthen feature interaction, at the cost of increased complexity. 
CANet~\cite{ren2021salient} addressed complex environments with a context-aware attention mechanism that models both global and local dependencies, while Wu \etal~\cite{wu2022edn} relied on extreme downsampling to capture global context, subsequently reconstructing fine-grained details through a Scale-Correlated Pyramid Convolution module. Nevertheless, balancing global context aggregation and local detail preservation remains a significant ongoing challenge due to the inherent trade-off between context coverage and fine-detail retention.

Transformer-based architectures have recently emerged as powerful alternatives for supervised SOD, leveraging global self-attention mechanisms. LOST~\cite{simeoni2021localizing} expands salient regions from seed patches identified through Vision Transformers (ViT), while TokenCut~\cite{wang2022self} employs normalised cuts~\cite{shi2000normalized} on self-attention features from ViTs to segment salient objects effectively. Although transformer-based models achieve robust performance, especially in challenging scenarios, their high computational requirements and reliance on extensive annotated datasets limit their practical deployment, particularly in real-time or resource-constrained settings. Additionally, the complexity of transformer models poses interpretability and scalability challenges, prompting further research into developing lightweight and efficient variants.

To alleviate annotation costs, weakly supervised methods have been explored, relying on sparse labels such as image-level class annotations, captions, scribbles, and bounding boxes~\cite{zeng2019multi, zhang2020weakly, wang2017learning}. Image-level labels~\cite{wang2017learning} and bounding boxes~\cite{khoreva2017simple} have proven effective for foreground object identification but frequently yield ambiguous object boundaries and limited segmentation accuracy. Scribble annotations~\cite{zhang2020weakly} offer improved precision but still necessitate manual effort and may fail to comprehensively cover the target objects.

Unsupervised methods \cite{zhang2017supervision, zhang2017learning} seek to eliminate reliance on annotations entirely, primarily through the generation and refinement of pseudo-labels. Zeng \etal~\cite{zeng2019multi} developed fusion strategies that combine outputs from multiple unsupervised models to create supervisory signals. Nguyen \etal~\cite{nguyen2019deepusps} iteratively refined pseudo-labels originating from various handcrafted methods, while Zhang \etal~\cite{zhang2020learning} integrated saliency prediction networks with noise-modelling modules for joint learning from noisy pseudo-labels. However, unsupervised methods frequently encounter instability in pseudo-label quality, constraining their overall accuracy and robustness.

Despite substantial progress, existing supervised and transformer-based models remain heavily dependent on extensive annotation, weakly supervised methods compromise precision, and unsupervised methods struggle with noisy label generation. To address these limitations, we introduce POTNet, an unsupervised framework integrating a novel hybrid clustering approach within the Prototypical Optimal Transport (POT) \cite{Wang2025POT} paradigm, where Wang et al.~\cite{Wang2025POT} proposed a weakly supervised segmentation framework that uses K-means-based class prototypes and optimal transport to refine pseudo labels via set-to-set matching, addressing the sparsity of CAMs. However, their approach relies solely on K-means and does not exploit complementary clustering strategies for adaptively grouping regions based on scene complexity and structure. In POTNet, we adaptively combine spectral and k-means clustering and generate semantically coherent pseudo masks without annotations. These high-quality masks subsequently guide the transformer-based AutoSOD network, enabling annotation-free, accurate, and robust salient object detection.

\begin{figure*}[t]
    \centering
    \includegraphics[width=\linewidth]{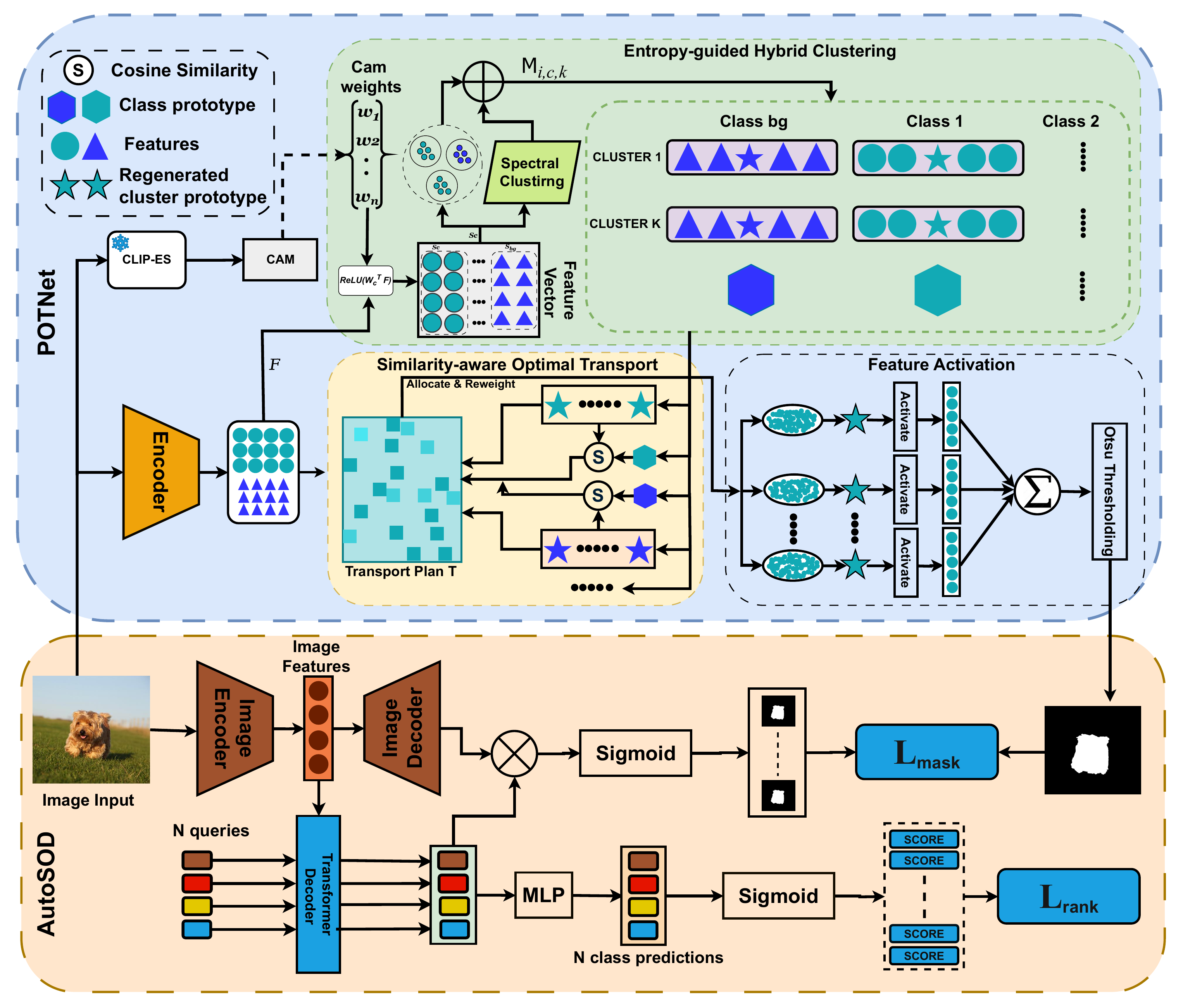}
        \caption{Overview of the proposed POTNet framework (top). It consists of prototype generation via entropy guided hybrid clustering, similarity-aware feature-to-prototype assignment using Optimal Transport, and prototype reweighting to produce pseudo-saliency masks, which supervise the AutoSOD network (bottom) for final prediction.}
    \label{fig:architecture}
\end{figure*}

\section{Annotation-Free Salient Object Detection}\label{sec:method}

Our pipeline (Fig.~\ref{fig:architecture}) converts self-supervised image features into reliable pseudo-saliency masks and trains a segmentation network without human labels. The proposed architecture \textbf{POTNet} realises our three core modules:  
(i) \emph{entropy-guided hybrid clustering} that yields semantically coherent prototypes,  
(ii) \emph{similarity-aware optimal transport} (OT) for feature-to-prototype assignment, and  
(iii) a \emph{confidence-weighted consistency loss} that refines prototype CAMs.  
The resulting masks supervise a transformer decoder, \textbf{AutoSOD}, producing final saliency maps.

\subsection{Problem Formulation}\label{subsec:prob_formulation}

Given an RGB image $\mathbf{x}\in\mathbb{R}^{H\times W\times3}$, unsupervised SOD seeks a mapping $f_{\theta}$ that partitions pixels into foreground (1) and background (0):
\begin{equation}
f_{\theta}(\mathbf{x}) = \hat{\mathbf{y}}\;\in\;\left\{0,1\right\}^{H\times W},
\end{equation}
where $\theta$ denotes learnable parameters. No ground-truth mask is available; supervision is provided solely by the pseudo-masks generated by POTNet. $\mathcal{C}$ denotes the set of foreground classes, $|\mathcal{C}|$ its cardinality, 
and $K$ is the number of prototypes per class. $\mathbf{1}_n$ is an $n$-dimensional vector of ones, $\mathbf{1}_{\{\cdot\}}$ is an indicator function. The helper function $\mathbf{g}(\mathbf{F},i)$ extracts a multi-level feature vector at pixel~$i$. POTNet transforms feature maps into high-quality saliency masks via three stages: prototype generation, similarity-aware OT assignment, and prototype re-weighting with a confidence loss.

\subsection{POTNet: Hybrid-Clustering Prototype Generation}\label{subsec:potnet}

\paragraph{Feature backbone and CAMs}
Let $\mathbf{F}\in\mathbb{R}^{D\times H\times W}$ be the last-layer features of a self-supervised encoder and $\mathbf{w}_c\in\mathbb{R}^{D}$ the linear classifier weight for class $c\in\mathcal{C}$. Class-activation maps (CAMs) are
\begin{equation}
\mathbf{S}_c = \operatorname{ReLU}\!\bigl(\mathbf{w}_c^{\!\top}\!\mathbf{F}\bigr),\quad
\mathbf{S}_{\text{bg}} = 1-\max_{c\in\mathcal{C}}\mathbf{S}_c.
\end{equation}

\paragraph{Entropy-guided hybrid clustering}

For each class $c$ (including $\text{bg}$), we select pixel vectors whose CAM exceeds a threshold $\tau$. We compute pixel-wise CAM entropy \ak{$\eta_i = -\sum_{c\in\mathcal C\cup\{\text{bg}\}} P_{i,c}\log P_{i,c},\; P_{i,c}=\frac{S_c(i)}{\sum_{c’}S_{c’}(i)}$}.  High-entropy pixels indicate global ambiguity, whereas low-entropy pixels are locally consistent. Pixels with $\eta_i\ge\eta_{\text{th}}$ are clustered by \emph{spectral} clustering to capture global structure, the remainder use \emph{k-means} for local compactness. Soft cluster assignment for pixel $i$ and prototype $k$ is 

\ak{
\begin{equation}
   M_{i,c,k} \;=\;
\begin{cases}
\,C^{\mathrm{spc}}_{i,c,k}, & \text{if } \eta_i \;\ge\; \eta_{\mathrm{th}},\\[4pt]
\,C^{\mathrm{km}}_{i,c,k},  & \text{otherwise}.
\end{cases}
\label{eq:hard_switch}
\end{equation}
}
\ak{where $C^{\mathrm{spc}}$  (resp. $C^{\mathrm{km}}$)  is the soft assignment returned by spectral (resp. k-means) clustering and $\eta_{\mathrm{th}}$ is a fixed threshold that separates globally ambiguous from locally compact pixels. For numerical stability, we replace the hard switch (Eq. \ref{eq:hard_switch}) with a
\textit{smooth entropy-gated fusion}
\begin{align}
  \gamma_i &= \frac{\eta_i-\eta_{\min}}
                  {\eta_{\max}-\eta_{\min}+\varepsilon},\qquad
  0\le\gamma_i\le 1, \tag{4}\\[2pt]
  M_{i,c,k} &=
       \gamma_i\,C^{\mathrm{spc}}_{i,c,k}
     + (1-\gamma_i)\,C^{\mathrm{km}}_{i,c,k}, \tag{5}
\end{align}
where $\eta_{\min}$ and $\eta_{\max}$ are the minimum and maximum
entropies among pixels whose class-$c$ CAM exceeds the threshold $\tau$. A very small value $\varepsilon$ is added to avoid division by zero. The gate $\gamma_i$ continuously biases high-entropy pixels toward the
spectral assignment (first term) and low-entropy pixels toward the k-means
assignment (second term), yielding a smooth \textit{entropy-guided hybrid
clustering}.  The resulting membership tensor $M$ is subsequently fed to
the Optimal-Transport stage
The memberships $M$ are used only to compute the prototypes $\mathbf{Z}$, which are then fed to the Optimal-Transport stage.} 

\paragraph{Prototype computation.}
Given $M\in [0,1]^{HW\times (|\mathcal{C}|+1)\times K}$, the $k$-th prototype of class $c$ is the spatial average of its assigned features:
\begin{equation}
\mathbf{Z}_{c,k}= \frac{\sum_{i} M_{i,c,k}\,\mathbf{g}(\mathbf{F},i)}{\sum_{i} M_{i,c,k}},
\end{equation}
yielding $\mathbf{Z}\in\mathbb{R}^{(|\mathcal{C}|+1)\times K\times D}$ capturing intra-class diversity. \ak{Intuitively, each row of $\mathbf{Z}$ now represents a semantic part prototype (e.g. head, torso, tail) for class $c$.}


\subsection{Similarity-Aware Optimal Transport Assignment}
\label{subsec:ot_assignment}
Following the prototype generation process, the next step involves assigning each feature vector to its most appropriate cluster prototype. A significant challenge at this stage is the accurate measurement of the distance between feature vectors and cluster prototypes. Conventional approaches, such as cosine similarity, evaluate the distance between individual feature-prototype pairs and assign each feature to its nearest prototype. However, such pointwise methods overlook inter-feature dependencies and treat assignments in isolation, often leading to fragmented and noisy segmentation masks.

To overcome this limitation, we reformulate feature allocation as a set-to-set problem and leverage Optimal Transport (OT) theory. OT provides a principled approach to aligning two distributions—here, the feature distribution and the prototype distribution—under global consistency constraints. By incorporating both similarity scores and distributional structure, our approach achieves more semantically meaningful feature assignments, improving the quality of generated saliency masks.

\paragraph{Feature and prototype sets.}
Let $\mathcal{F}=\{\mathbf{f}_i\}_{i=1}^{N}$ denote the $D$-dimensional pixel features ($N=H \times W$), and $\mathcal{Z}=\{\mathbf{Z}_{c,k}\}_{k=1}^{K}$ the prototypes of class $c \in \mathcal{C}$ from Sec.~\ref{subsec:potnet}. To quantify feature-prototype dissimilarity, we define the cost matrix
\[
\mathbf{C}_{i,j}=1-\frac{\mathbf{f}_i^{\!\top}\mathbf{Z}_j}{\lVert\mathbf{f}_i\rVert\,\lVert\mathbf{Z}_j\rVert},\qquad \mathbf{C}_{i,j}\in [0,2],
\]
and derive the corresponding similarity matrix as $\mathbf{S}=1-\mathbf{C}$.

\paragraph{Prototype-aware column marginals.}
To guide feature assignments towards semantically meaningful prototypes, we modulate the prototype importance using classifier similarity. Specifically, for each prototype $\mathbf{Z}_{c,k}$, we compute its cosine similarity to the corresponding classifier weight $\mathbf{w}_c$:
\[
s_{c,k}= \frac{\mathbf{Z}_{c,k}^{\!\top}\mathbf{w}_c}{\lVert\mathbf{Z}_{c,k}\rVert\,\lVert\mathbf{w}_c\rVert}.
\]
These similarities are normalised via a softmax function:
\[
\mathbf{v} = \operatorname{softmax}\bigl(\mathrm{vec}(s_{c,k})\bigr)\in\mathbb{R}^{K(|\mathcal{C}|+1)},
\]
producing adaptive column marginals that assign higher importance to prototypes aligned with classifier semantics.

\paragraph{OT formulation.}
The objective of the OT assignment is to compute a transport plan $\mathbf{T}\in\mathbb{R}_{+}^{N \times P}$ that maximises overall feature-prototype similarity while satisfying balanced marginal constraints. Formally, we solve
\begin{equation}
\mathbf{T}^{*} = \arg\max_{\mathbf{T}}\;\langle\mathbf{T},\mathbf{S}\rangle
\quad \text{s.t.} \quad \mathbf{T}\mathbf{1}_P = \tfrac{1}{N}\mathbf{1}_N, \quad \mathbf{T}^{\!\top}\mathbf{1}_N = \mathbf{v},
\end{equation}
where $P = K(|\mathcal{C}|+1)$ denotes the total number of prototypes. The first constraint ensures equal weighting across all features, while the second enforces prototype importance based on classifier-guided marginals. We employ Sinkhorn iterations~\cite{chizat2018scaling} for efficient computation.

\paragraph{Hard assignment.}
After obtaining the optimal transport matrix $\mathbf{T}^{*}$, we derive discrete assignments by selecting the most probable prototype for each pixel:
\[
j^{*}(i)=\arg\max_{j}\,\mathbf{T}^{*}_{i,j}, \quad \widehat{M}_{i,j}= \mathbf{1}_{\{j=j^{*}(i)\}}.
\]

The proposed OT assignment strategy introduces a global, structure-aware approach to feature clustering. By integrating classifier-informed prototype marginals and globally consistent optimal transport matching, this mechanism ensures that feature assignments are both semantically meaningful and distributionally balanced. Unlike conventional nearest-neighbour approaches, our method prevents prototype under-utilisation and avoids noisy or fragmented assignments. This stage plays a pivotal role in producing clean and accurate pseudo-saliency masks, providing a robust supervisory signal for subsequent saliency prediction in AutoSOD.

\subsection{Prototype Re-weighting \& Confidence-Weighted Consistency Loss}
\label{subsec:reweight}
This section exploits the OT plan to (i) update prototypes with newly
assigned features and (ii) weight the supervision signal by assignment
confidence, yielding sharper, noise-aware pseudo-masks.

\paragraph{Prototype update.}
Let $\widehat{M}\in\{0,1\}^{N\times P}$ be the hard assignment mask from
Sec.~\ref{subsec:ot_assignment}.  
We regenerate prototypes as class-conditioned averages of their assigned
features:
\begin{equation}
  \mathbf{Z}^{\dagger}_{c,k}
  \;=\;
  \frac{\sum_{i}\widehat{M}_{i,(c,k)}\,\mathbf{g}(\mathbf{F},i)}
       {\sum_{i}\widehat{M}_{i,(c,k)}}.
  \label{eq:proto_update}
\end{equation}

\paragraph{Prototype CAMs and OT re-weighting.}
The CAM of prototype $(c,k)$ is
\(
  \mathcal{M}_{c,k}= \operatorname{sim}\!\bigl(\mathbf{Z}^{\dagger}_{c,k},
  \mathbf{F}\bigr),
\)
where \texttt{sim} denotes cosine similarity.  
We re-weight it by the averaged OT mass
\(
  \gamma_{c,k}=\tfrac{1}{N}\sum_{i}\mathbf{T}^{*}_{i,(c,k)}
\)
to emphasise reliable prototypes:
\begin{equation}
  \widetilde{\mathcal{M}}_{c,k}= \gamma_{c,k}\,\mathcal{M}_{c,k}\,.
\end{equation}
The class-level CAM is the sum over its $K$ prototypes,
\(
  \widetilde{\mathcal{M}}_{c}= \sum_{k=1}^{K}\widetilde{\mathcal{M}}_{c,k}.
\)

\paragraph{Confidence-weighted consistency loss.}
We align the re-weighted prototype CAMs with the original classifier CAMs
$\mathcal{M}^{\text{cls}}$ via an $L_1$ loss modulated by pixel confidence
$\omega_{i}$ (low entropy $\Rightarrow$ high weight):
\begin{equation}
  \mathcal{L}_{\text{con}}
  =
  \frac{1}{N}
  \sum_{i=1}^{N}\!\omega_{i}
  \bigl\|
    \widetilde{\mathcal{M}}_{i}-\mathcal{M}^{\text{cls}}_{i}
  \bigr\|_{1},
  \qquad
  \omega_{i}=1-\mathcal{H}\!\bigl(\mathbf{T}^{*}_{i,\cdot}\bigr),
  \label{eq:consistency}
\end{equation}
where $\mathcal{H}(\cdot)$ is Shannon entropy normalised to $[0,1]$.

\paragraph{Pseudo-mask extraction.}
After convergence, we apply Otsu’s threshold to
$\widetilde{\mathcal{M}}_{\text{fg}}=\max_{c\neq\text{bg}}\widetilde{\mathcal{M}}_{c}$
to obtain the binary pseudo-mask
$\mathbf{M}^{\text{pseudo}}\in\{0,1\}^{H\times W}$ that supervises
AutoSOD.  All steps are annotation-free and differentiable, enabling
end-to-end training of POTNet.

Through prototype re-weighting and confidence-aware consistency loss, this stage refines the initially coarse saliency maps into sharper and more reliable pseudo labels. The use of OT-based confidence weighting ensures that high-certainty predictions contribute more significantly to the training objective, reducing the impact of noisy or ambiguous regions. 

\begin{table*}[t]
\centering
\scriptsize
\renewcommand{\arraystretch}{1.1}
\setlength{\tabcolsep}{3pt}
\caption{Comparison with SOTA methods on five benchmark datasets (\textbf{DUTS-TE}, \textbf{DUT-OMRON}, \textbf{HKU-IS}, \textbf{PASCAL}, and \textbf{ECSSD}) using the metrics $S_m$, $F_\beta$, $E_\eta$, and MAE where $\uparrow$ and $\downarrow$ denote larger and smaller is better, respectively.}
\label{tab:sod_comparison_fixed}

\begin{tabular}{|l|cccc|cccc|cccc|cccc|cccc|}
\hline
\multirow{2}{*}{\textbf{Method}} & \multicolumn{4}{c|}{\textbf{DUTS-TE}} & \multicolumn{4}{c|}{\textbf{DUT-OMRON}} & \multicolumn{4}{c|}{\textbf{HKU-IS}} & \multicolumn{4}{c|}{\textbf{PASCAL}} & \multicolumn{4}{c|}{\textbf{ECSSD}} \\
 & $S_m \uparrow$ & $F_\beta \uparrow$ & $E_\eta \uparrow$ & MAE $\downarrow$ & $S_m$ & $F_\beta$ & $E_\eta$ & MAE & $S_m$ & $F_\beta$ & $E_\eta$ & MAE & $S_m$ & $F_\beta$ & $E_\eta$ & MAE & $S_m$ & $F_\beta$ & $E_\eta$ & MAE \\
\hline
\multicolumn{21}{|c|}{\textbf{Fully Supervised SOD methods}} \\
\hline
PiCANet \cite{liu2018picanet} & 0.851 & 0.757 & 0.853 & 0.062 & 0.826 & 0.710 & 0.823 & 0.072 & 0.906 & 0.854 & 0.909 & 0.047 & 0.848 & 0.799 & 0.804 & 0.129 & 0.867 & 0.871 & 0.909 & 0.054 \\
MSNet \cite{wu2019mutual}   & 0.851 & 0.792 & 0.883 & 0.050 & 0.809 & 0.709 & 0.830 & 0.064 & 0.907 & 0.878 & 0.930 & 0.039 & 0.844 & 0.671 & 0.813 & 0.822 & 0.905 & 0.885 & 0.922 & 0.048 \\
BASNet \cite{qin2019basnet}  & 0.866 & 0.828 & 0.896 & 0.048 & 0.836 & 0.767 & 0.865 & 0.057 & 0.909 & 0.903 & 0.943 & 0.032 & 0.838 & 0.821 & 0.821 & 0.122 & 0.910 & 0.913 & 0.938 & 0.040 \\
U2Net \cite{qin2020u2}   & 0.861 & 0.807 & 0.889 & 0.051 & 0.857 & 0.773 & 0.867 & 0.054 & 0.916 & 0.910 & 0.936 & 0.033 & 0.844 & 0.797 & 0.831 & 0.074 & 0.918 & 0.910 & 0.936 & 0.033 \\
VST \cite{liu2021visual}     & 0.885 & 0.870 & 0.893 & 0.039 & 0.863 & 0.829 & 0.865 & 0.067 & 0.919 & 0.922 & 0.962 & 0.028 & 0.863 & 0.829 & 0.865 & 0.067 & 0.917 & 0.929 & 0.945 & 0.034 \\
\hline
\multicolumn{21}{|c|}{\textbf{Weakly Supervised SOD methods}} \\
\hline
WSS \cite{wang2017learning}     & 0.748 & 0.633 & 0.806 & 0.073 & 0.730 & 0.590 & 0.729 & 0.110 & 0.822 & 0.773 & 0.819 & 0.079 & - & 0.698 & 0.690 & 0.184 & 0.808 & 0.767 & 0.796 & 0.108 \\
WSI \cite{li2018weakly}     & 0.697 & 0.569 & 0.690 & 0.075 & 0.759 & 0.641 & 0.761 & 0.100 & 0.808 & 0.763 & 0.800 & 0.089 & - & 0.653 & 0.647 & 0.206 & 0.805 & 0.762 & 0.792 & 0.068 \\
MSW \cite{zeng2019multi}     & 0.759 & 0.648 & 0.742 & 0.076 & 0.756 & 0.597 & 0.728 & 0.109 & 0.818 & 0.734 & 0.786 & 0.084 & 0.697 & 0.685 & 0.693 & 0.178 & 0.825 & 0.761 & 0.787 & 0.098 \\
Scribble\_S \cite{zhang2020weakly} & 0.793 & 0.746 & 0.865 & 0.062 & 0.771 & 0.702 & 0.835 & 0.068 &
0.855 & 0.857 & 0.923 & 0.047 & 0.742 & 0.788 & 0.798 & 0.140 & 0.854 & 0.865 & 0.908 & 0.061 \\
MFNet \cite{piao2021mfnet}   & 0.775 & 0.770 & 0.620 & 0.762 & 0.074 & 0.803 & 0.646 & 0.846 & 0.087 & 0.851 & 0.921 & 0.059 & 0.770 & 0.751 & 0.817 & 0.115 & 0.834 & 0.854 & 0.885 & 0.084 \\
\hline
\multicolumn{21}{|c|}{\textbf{Unsupervised SOD methods}} \\
\hline
SBF \cite{zhang2017supervision}     & 0.739 & 0.612 & 0.763 & 0.108 & 0.731 & 0.612 & 0.763 & 0.108 & 0.812 & 0.783 & 0.855 & 0.075 & 0.712 & 0.735 & 0.746 & 0.167 & 0.813 & 0.782 & 0.835 & 0.096 \\
MNL \cite{zhang2017learning}     & 0.813 & 0.725 & 0.853 & 0.075 & 0.733 & 0.597 & 0.712 & 0.103 & 0.860 & 0.820 & 0.858 & 0.065 & 0.728 & 0.748 & 0.741 & 0.158 & 0.845 & 0.810 & 0.836 & 0.090 \\
EDNS \cite{zhang2020learning}    & 0.828 & 0.747 & 0.859 & 0.060 & 0.791 & 0.701 & 0.816 & 0.070 & 0.890 & \textbf{0.878} & 0.919 & 0.043 & 0.750 & 0.759 & 0.794 & 0.142 & 0.860 & 0.852 & 0.883 & 0.071 \\
DeepUSPS \cite{nguyen2019deepusps}& 0.787 & 0.734 & 0.848 & 0.068 & 0.795 & 0.713 & 0.848 & 0.063 & 0.876 & 0.864 & 0.930 & 0.041 & 0.757 & 0.768 & 0.792 & 0.151 & 0.861 & 0.870 & 0.903 & 0.063 \\
Yan et al. \cite{yan2022unsupervised}& 0.840 & 0.772 & 0.859 & 0.052 & 0.801 & 0.711 & 0.841 & 0.066 & 0.890 & 0.873 & 0.931 & 0.047 & 0.759 & 0.770 & 0.792 & 0.158 & 0.862 & \textbf{0.876} & 0.888 & 0.068 \\
UMNet \cite{wang2022multi}   & 0.802 & 0.749 & 0.863 & 0.067 & 0.804 & 0.727 & 0.859 & 0.063 & 0.886 & 0.872 & 0.939 & 0.041 & 0.762 & 0.775 & 0.800 & 0.144 & 0.867 & 0.872 & 0.902 & 0.064 \\
\textbf{AutoSOD(ours)}   & \textbf{0.869} & \textbf{0.776} & \textbf{0.877} & \textbf{0.042} & \textbf{0.834} & \textbf{0.763} & \textbf{0.863} & \textbf{0.059} & \textbf{0.913} & 0.870 & \textbf{0.947} & \textbf{0.038} & \textbf{0.783} & \textbf{0.859} & \textbf{0.817} & \textbf{0.119} & \textbf{0.912} & 0.843 & \textbf{0.928} & \textbf{0.037} \\
\hline
\end{tabular}
\end{table*}

\begin{figure*}[ht]
    \centering
    \setlength{\tabcolsep}{1pt}
    \renewcommand{\arraystretch}{1}
    \begin{tabular}{ccccccccccc}
        \textbf{Image} & \textbf{GT} & \textbf{LDF(S)} & \textbf{BASNet(S)} & \textbf{U2Net(S)} & \textbf{Scribble\_S(W)} &  \textbf{EDNS(U)} & \textbf{AutoSOD(ours)} \\
        
        \includegraphics[width=0.11\textwidth]{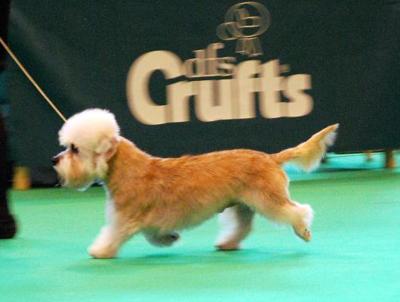} &
        \includegraphics[width=0.11\textwidth]{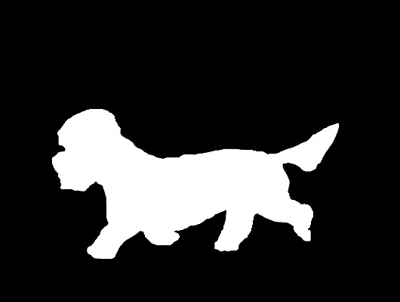} &
        \includegraphics[width=0.11\textwidth]{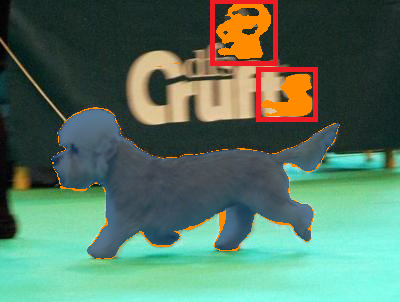} &
        \includegraphics[width=0.11\textwidth]{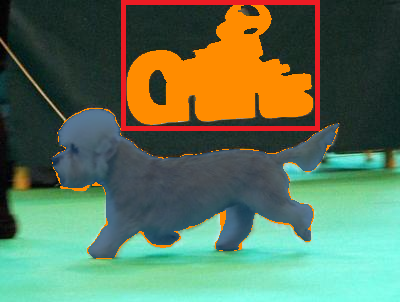} &
        \includegraphics[width=0.11\textwidth]{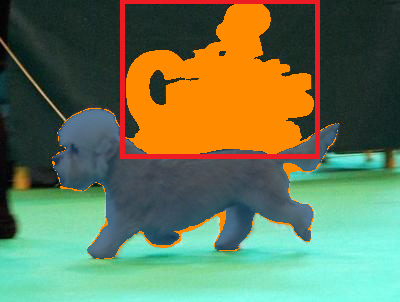} &
        \includegraphics[width=0.11\textwidth]{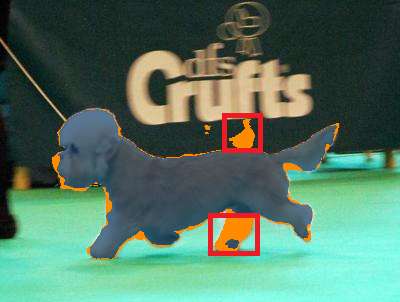} &
        
        \includegraphics[width=0.11\textwidth]{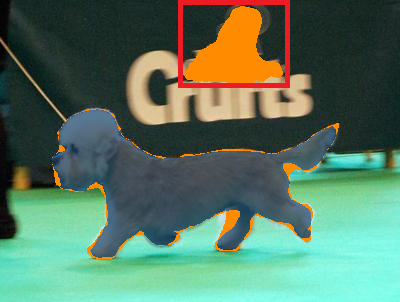} &
        \includegraphics[width=0.11\textwidth]{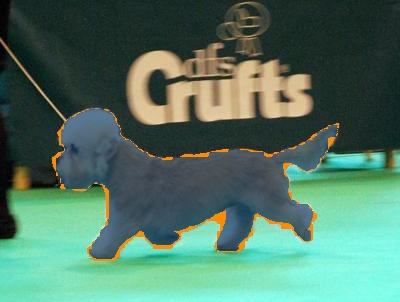}
        \\
        
        \includegraphics[width=0.11\textwidth]{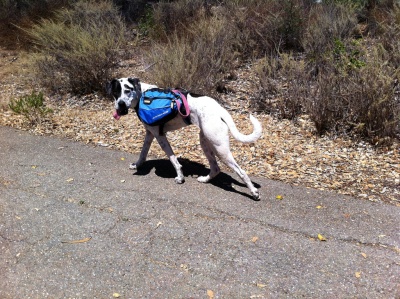} &
        \includegraphics[width=0.11\textwidth]{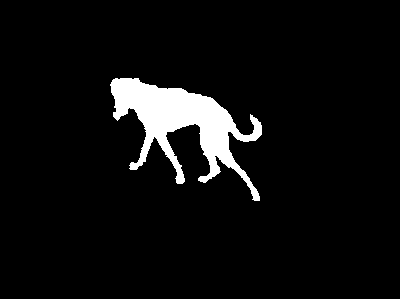} &
        \includegraphics[width=0.11\textwidth]{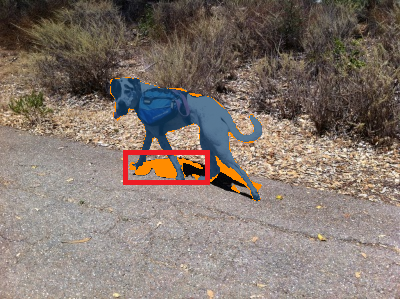} &
        \includegraphics[width=0.11\textwidth]{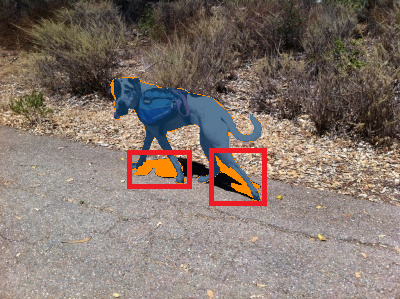} &
        \includegraphics[width=0.11\textwidth]{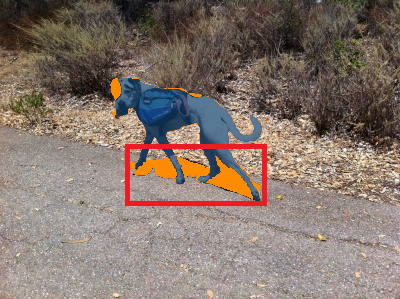} &
        \includegraphics[width=0.11\textwidth]{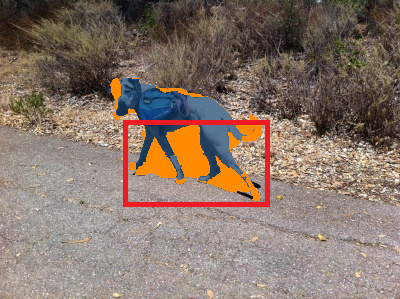} &
        
        \includegraphics[width=0.11\textwidth]{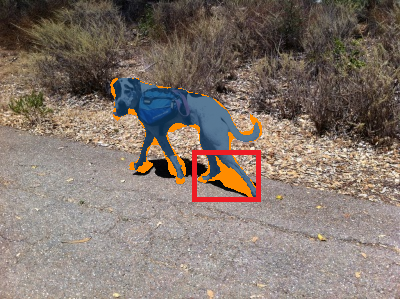} &
        \includegraphics[width=0.11\textwidth]{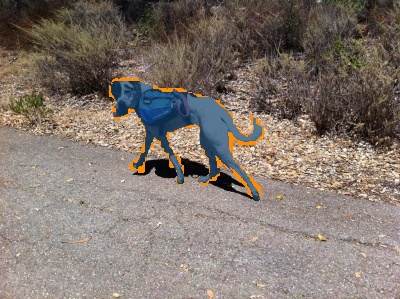} \\

        \includegraphics[width=0.11\textwidth]{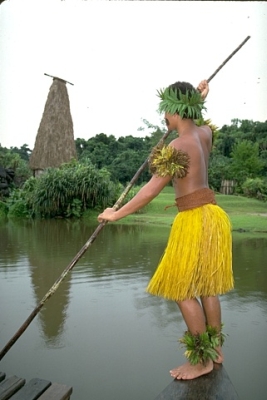} &
        \includegraphics[width=0.11\textwidth]{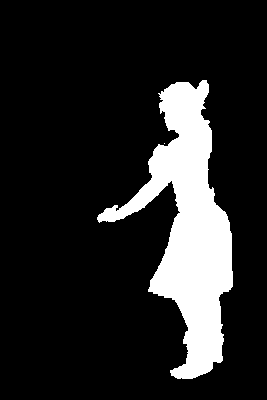} &
        \includegraphics[width=0.11\textwidth]{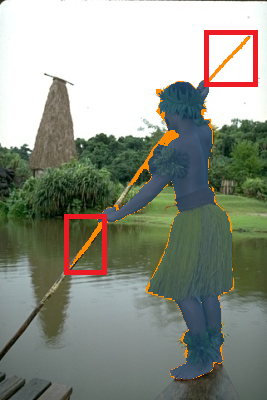} &
        \includegraphics[width=0.11\textwidth]{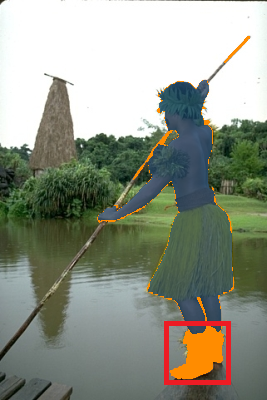} &
        \includegraphics[width=0.11\textwidth]{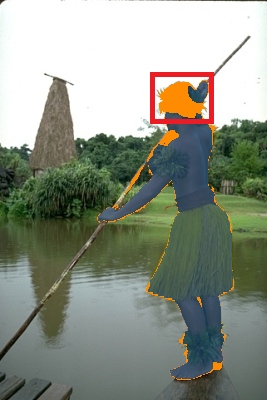} &
        \includegraphics[width=0.11\textwidth]{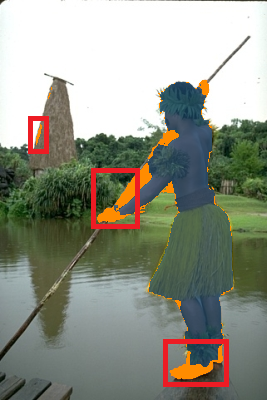} &
        
        \includegraphics[width=0.11\textwidth]{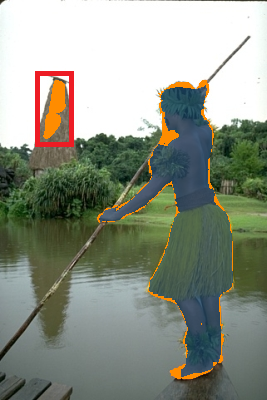} &
        \includegraphics[width=0.11\textwidth]{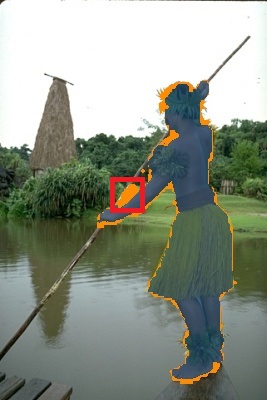} \\
    \end{tabular}
    \caption{Qualitative comparison of saliency maps predicted by different methods. The \textbf{S}, \textbf{W}, and \textbf{U} show the supervised, weakly supervised and unsupervised domains of each method, respectively. Each row shows: input image, ground truth (GT), and predictions from fully supervised methods (LDF, BASNet, U2Net), weakly supervised (Scribble\_S), unsupervised (EDNS) methods, and AutoSOD(ours). The red boxes indicate the error in the saliency predictions by different methods. 
    }
    \label{fig:saliency-comparison}
\end{figure*}

\subsection{Automatic Salient Object Detection (AutoSOD) Network}\label{subsec:AutoSOD}


AutoSOD is built upon a modified version of the MaskFormer architecture~\cite{cheng2021per}, originally designed for semantic and instance segmentation tasks. Our model generates two outputs: binary saliency masks and corresponding objectness scores for each mask.

The architecture consists of three key components: an image encoder, a pixel decoder, and a transformer decoder. Given an input image $\mathbf{x} \in \mathbb{R}^{H \times W \times 3}$, the encoder extracts features $\phi(\mathbf{x}) \in \mathbb{R}^{h \times w \times D}$. These features are then passed through a pixel decoder, producing an upsampled feature map $\psi$.  

Simultaneously, the transformer decoder receives the encoded features as keys and values, and a set of learnable queries as input. It generates $N_q$ per-mask embeddings. Final mask predictions $\mathbf{M} = \{\mathbf{M}_i\}_{i=1}^{N_q}$ are obtained by taking the dot product of the upsampled feature map and each per-mask query embedding $\mathbf{q}_i \in \mathbb{R}^{D}$, followed by a sigmoid activation. 


An objectness score $o_i \in [0,1]$ for each mask $\mathbf{M}_i$ is predicted by feeding the corresponding embedding $\mathbf{q}_i$ into a two-layer MLP followed by a sigmoid function. Note that this framework is flexible and allows alternative encoder-decoder configurations.

To train AutoSOD, we use a combination of two loss functions: a mask similarity loss $\mathcal{L}_{\text{mask}}$ and a ranking loss $\mathcal{L}_{\text{rank}}$.

We utilise the Dice coefficient 
to define the mask loss, which effectively handles the class imbalance between foreground and background regions. Let $\mathbf{M}^{\text{pseudo}}$ denote the generated pseudo-mask. The Dice loss is computed between each predicted mask and the pseudo-mask.

To select the most accurate mask when $N_q > 1$, we impose a ranking constraint based on the objectness score. Specifically, we first sort the predicted masks by their mask loss in ascending order. 
To enforce consistency between mask quality and objectness, we apply a hinge-based ranking loss:

\begin{equation}
    \mathcal{L}_{\text{rank}} = \sum_{i=1}^{N_q - 1} \sum_{j > i}^{N_q} \max(0,~o_j - o_i).
\end{equation}

The complete loss function is given by:

\begin{equation}
    \mathcal{L} = \mathcal{L}_{\text{mask}} + \lambda \mathcal{L}_{\text{rank}},
\end{equation}

where $\lambda$ is a balancing hyperparameter, set to 1.0 in all experiments. Following common practice,
we compute the loss at each transformer decoder layer.

During inference, we select the predicted mask with the highest objectness score among the $N_q$ predictions. This mask is then binarised using a fixed threshold of 0.5 to produce the final saliency map.

\section{Experiments and Results}\label{sec:results and discussion}

\subsection{Datasets}\label{datasets}



We use the DUTS-TR dataset, which consists of 10,553 images, to train our model using the pseudo-masks. It is important to note that only the raw images are used for both pseudo-mask generation and training, without relying on any ground-truth annotations.

For ablation studies and comparisons with prior work, we evaluate our method on five standard saliency detection datasets: 
\textit{(i)}~DUT-OMRON, which contains 5,168 images with diverse scenes and pixel-level annotations; 
\textit{(ii)}~DUTS-TE, comprising 5,019 test images; 
\textit{(iii)}~ECSSD, which includes 1,000 images with complex foreground-background structures; 
\textit{(iv)}~HKU-IS, consisting of 4,447 images where foreground and background often appear visually similar; and 
\textit{(v)}~PASCAL-S, with 850 images featuring multiple salient objects.

\subsection{Implementation Details}\label{implementation_details}
We adopt the ViT-S/8 architecture~\cite{dosovitskiy2020image} as the encoder in our model. For the pixel decoder, we employ a simple bilinear upsampling operation with a scale factor of 2. The transformer decoder comprises six layers, following the design in~\cite{vaswani2017attention}.

To predict objectness scores from the per-mask embeddings of size 384, we use a multi-layer perceptron (MLP) consisting of three fully connected layers. A ReLU activation function is applied between the layers. The number of hidden units is kept equal to the input dimensionality, and the final output is a single scalar value passed through a sigmoid function to produce the objectness probability.

\subsection{Results and Discussion}\label{results}

\subsubsection{Quantitative Results}
We evaluate the performance of our proposed method, AutoSOD, against different SOTA weakly supervised, unsupervised, and fully supervised methods as shown in Table \ref{tab:sod_comparison_fixed}. For a comprehensive evaluation, we employ commonly used metrics including structure measure ($S_m$), mean F-score ($F_{\beta}$), mean E-measure ($E_{\eta}$), and mean absolute error (MAE).
As shown in Table~\ref{tab:sod_comparison_fixed}, our AutoSOD model consistently surpasses state-of-the-art weakly and unsupervised methods across all datasets and evaluation metrics. This demonstrates the effectiveness of our POTNet, a pseudo-ground truth generation strategy, compared to existing weakly or unsupervised SOD techniques. Notably, although AutoSOD is trained without any human annotations or weak labels,
its performance remains competitive with fully supervised approaches.

\subsubsection{Qualitative Results}
To further demonstrate the effectiveness of AutoSOD, we provide qualitative comparisons with existing SOD methods, as illustrated in Fig.~\ref{fig:saliency-comparison}. In the first and second rows, where the salient objects are dogs in varying backgrounds, most methods exhibit minor artefacts, such as background logo or shadow or boundary imprecision. However, AutoSOD consistently produces cleaner, more accurate masks with well-preserved contours and fewer false positives.

The third row presents a significantly more challenging scene involving a person holding a thin stick, with water reflections and a cluttered background. Here, supervised methods like BASNet and U2Net miss the body parts of the person, including the head and foot. The weakly supervised approach Scribble$_S$ also misses the foot of the person. While the unsupervised method EDNS highlights irrelevant background regions. In contrast, AutoSOD accurately captures the entire object, including the irregular boundaries of the person, while effectively suppressing complex background information, demonstrating its superior capability in fine-structure preservation and complex region discrimination.

\begin{table}[ht]
\centering
\caption{Ablation study of AutoSOD on the DUTS-TE dataset.}
\begin{tabular}{lcccc}
\toprule
\textbf{Model Variant} & $S_m \uparrow$ & $F_\beta \uparrow$ & $E_\eta \uparrow$ & MAE $\downarrow$ \\
\midrule
Full (AutoSOD)          & \textbf{0.869} & \textbf{0.776} & \textbf{0.877} & \textbf{0.042} \\
Only K-means Clustering & 0.855          & 0.751          & 0.861          & 0.051 \\
Only Spectral Clustering & 0.857         & 0.754          & 0.863          & 0.049 \\
w/o OT Consistency Loss  & 0.849         & 0.743          & 0.856          & 0.053 \\
w/o Reweighting          & 0.842         & 0.737          & 0.848          & 0.055 \\
\bottomrule
\end{tabular}
\label{tab:ablation}
\end{table}

\subsection{Ablation Study}\label{ablation}

To evaluate the effectiveness of individual components within AutoSOD, we conduct an ablation study on the DUTS-TE dataset, as shown in Table~\ref{tab:ablation}. The full model achieves the best performance across all metrics, validating the benefit of the complete design. Replacing our hybrid clustering module with a single clustering method results in noticeable performance drops: the "Only K-means Clustering" variant shows limitations in capturing global semantic structures, while the "Only Spectral Clustering" variant lacks local compactness in feature space. These results support the idea that our hybrid strategy effectively balances global coherence and local discrimination. Removing the OT-based consistency loss further degrades results, indicating its essential role in aligning feature assignment with the initial CAM distribution. Likewise, removing the prototype reweighting mechanism reduces boundary precision and leads to less refined saliency maps. Altogether, the ablation confirms that each module, particularly the hybrid clustering design, plays a vital role in boosting saliency detection under unsupervised settings.

\section{Conclusion and Future Work}\label{sec:Conclusion}
We introduced AutoSOD, an end-to-end unsupervised salient object detection framework that leverages POTNet, a hybrid clustering mechanism guided by entropy and aligned via optimal transport. By combining spectral and k-means clustering based on spatial uncertainty, POTNet produces sharper, part-aware pseudo masks without handcrafted priors. These masks supervise a standard segmentation model, eliminating the need for manual annotations or offline post-processing. Extensive experiments demonstrate that AutoSOD sets a new state of the art in unsupervised SOD and further closes the gap with fully supervised approaches. In future work, we aim to investigate integrating temporal consistency for video SOD and explore more advanced clustering strategies that adapt dynamically across diverse scene types. Moreover, combining our approach with generative priors or diffusion-based refinement may further improve boundary precision and robustness in complex backgrounds.
\bibliographystyle{ieeetr} 
\bibliography{main}
\end{document}